\documentclass[manuscript, final, acmsmall]{acmart}

\AtBeginDocument{%
  }

\setcopyright{acmlicensed}
\copyrightyear{2025}
\acmYear{2025}
\acmDOI{XXXXXXX.XXXXXXX}

\acmJournal{TOMM}
\acmVolume{37}
\acmNumber{4}
\acmArticle{111}
\acmMonth{4}

\usepackage{hyperref}
\usepackage{bm}
\usepackage{mathrsfs}
\usepackage{amsmath}
\usepackage{subcaption}
\usepackage{colortbl}
\usepackage{graphicx}
\usepackage{multirow}                
\usepackage{booktabs}
\usepackage{caption}
\usepackage{xcolor} 
\usepackage[table]{xcolor} 
\hypersetup{
    colorlinks,
    linkcolor={blue},
    citecolor={blue},
    urlcolor={red}
}

\begin{document}

\title{Exploring Talking Head Models With Adjacent Frame Prior for Speech-Preserving Facial Expression Manipulation}





\author{Zhenxuan Lu}
\authornotemark[2]
\affiliation{%
  \institution{Guangdong University of Technology}
  \city{Guangzhou}
  \country{China}}
\email{2112303213@mail2.gdut.edu.cn}

\author{Zhihua Xu}
\authornotemark[2]
\affiliation{%
  \institution{Guangdong University of Technology}
  \city{Guangzhou}
  \country{China}}
\email{zihua@mail2.gdut.edu.cn}

\author{Zhijing Yang}
\authornotemark[1]
\affiliation{%
  \institution{Guangdong University of Technology}
  \city{Guangzhou}
  \country{China}}
\email{yzhj@gdut.edu.cn}

\author{Feng Gao}
\authornotemark[1]
\affiliation{%
  \institution{Peking University}
  \city{Beijing}
  \country{China}}
\email{gaof@pku.edu.cn}

\author{Yongyi Lu}
\affiliation{%
  \institution{Guangdong University of Technology}
  \city{Guangzhou}
  \country{China}}
\email{yylu1989@gmail.com}

\author{Keze Wang}
\affiliation{
  \institution{Sun Yat-sen University}
  \city{Guangzhou}
  \country{China}}
\email{kezewang@gmail.com}

\author{Tianshui Chen}
\affiliation{
  \institution{Guangdong University of Technology}
  \city{Guangzhou}
  \country{China}}
\email{tianshuichen@gmail.com}

\footnotetext[2]{Zhenxuan Lu and Zhihua Xu Contribute equally to this work and share co-first authorship.}
\footnotetext[1]{Zhijing Yang and Feng Gao are corresponding authors.}
\renewcommand{\shortauthors}{Zhenxuan Lu et al.}


\begin{abstract}
Speech-Preserving Facial Expression Manipulation (SPFEM) is an innovative technique aimed at altering facial expressions in images and videos while retaining the original mouth movements. Despite advancements, SPFEM still struggles with accurate lip synchronization due to the complex interplay between facial expressions and mouth shapes. Capitalizing on the advanced capabilities of audio-driven talking head generation (AD-THG) models in synthesizing precise lip movements, our research introduces a novel integration of these models with SPFEM. We present a new framework, Talking Head Facial Expression Manipulation (THFEM), which utilizes AD-THG models to generate frames with accurately synchronized lip movements from audio inputs and SPFEM-altered images. However, increasing the number of frames generated by AD-THG models tends to compromise the realism and expression fidelity of the images. To counter this, we develop an adjacent frame learning strategy that finetunes AD-THG models to predict sequences of consecutive frames. This strategy enables the models to incorporate information from neighboring frames, significantly improving image quality during testing. Our extensive experimental evaluations demonstrate that this framework effectively preserves mouth shapes during expression manipulations, highlighting the substantial benefits of integrating AD-THG with SPFEM.
\end{abstract}



\maketitle

\section{Introduction}
Speech-Preserving Facial Expression Manipulation (SPFEM) is a technique designed to modify facial expressions while preserving the original mouth movements in images or videos. This approach significantly enhances human expression, offering substantial benefits for virtual digital characters and film \& television production. For instance, capturing the precise emotions of actors in films typically requires significant effort and repetitive takes. In contrast, a robust SPFEM system can effortlessly modify facial expressions during post-production, achieving the desired emotive effect with greater efficiency. This capability not only fosters innovation in film and television production but also enables the creation of diverse facial expressions, underscoring the practical importance of SPFEM research.

Recent advancements in SPFEM have primarily focused on optimizing face reconstruction algorithms \cite{1,2,3,50,52,53} and decoupling semantic representations to facilitate expression editing \cite{4,5,8,54}. These improvements have allowed for more precise manipulations and better-quality rendering of facial expressions in various applications. However, a significant challenge still persists due to the intrinsic link between mouth shapes and facial expressions. SPFEM models struggle to disentangle these interconnected features. Consequently, expression editing often inadvertently alters the mouth shape, thus compromising the integrity of lip movements corresponding to spoken content.

Building upon these challenges, the central idea of this work is to combine the strengths of AD-THG models and SPFEM models. While SPFEM models are effective at editing and enriching facial expressions, AD-THG models particularly excel at producing accurate and synchronized lip movements that align with speech. By integrating these two complementary approaches, we achieve a framework that not only enables diverse and expressive facial manipulations but also preserves natural lip synchronization. Compared with relying on either SPFEM or AD-THG alone, this joint design markedly improves overall performance in both emotional expressiveness and lip-speech consistency.

\begin{figure}[h]
    \centering
    \begin{minipage}[b]{0.13\textwidth}
        \centering
        \includegraphics[width=\textwidth]{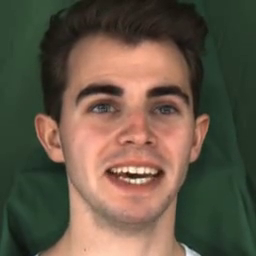}
    \end{minipage}
    \hspace{-1.3mm}
    \begin{minipage}[b]{0.13\textwidth}
        \centering
        \includegraphics[width=\textwidth]{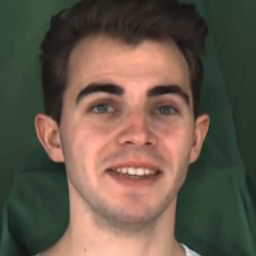}
    \end{minipage}
    \hspace{-1.3mm}
    \begin{minipage}[b]{0.13\textwidth}
        \centering
        \includegraphics[width=\textwidth]{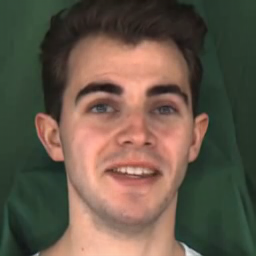}
    \end{minipage}
    \hspace{-1.3mm}
    \begin{minipage}[b]{0.13\textwidth}
        \centering
        \includegraphics[width=\textwidth]{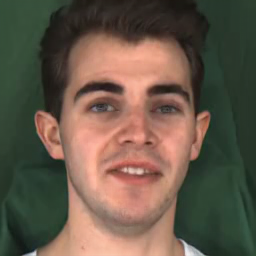}
    \end{minipage}

    \begin{minipage}[b]{0.13\textwidth}
        \centering
        \includegraphics[width=\textwidth]{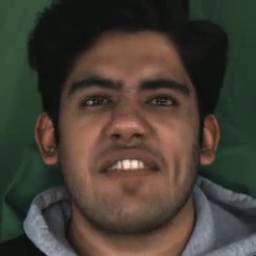}
    \end{minipage}
    \hspace{-1.3mm}
    \begin{minipage}[b]{0.13\textwidth}
        \centering
        \includegraphics[width=\textwidth]{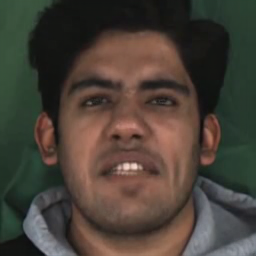}
    \end{minipage}
    \hspace{-1.3mm}
    \begin{minipage}[b]{0.13\textwidth}
        \centering
        \includegraphics[width=\textwidth]{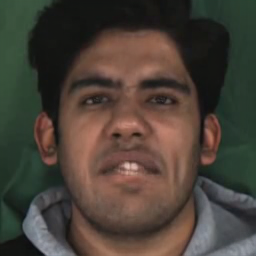}
    \end{minipage}
    \hspace{-1.3mm}
    \begin{minipage}[b]{0.13\textwidth}
        \centering
        \includegraphics[width=\textwidth]{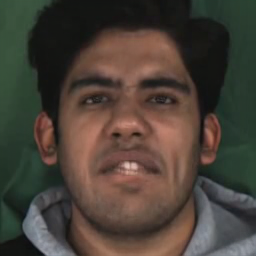}
    \end{minipage}

    \begin{minipage}[b]{0.13\textwidth}
        \centering
        \includegraphics[width=\textwidth]{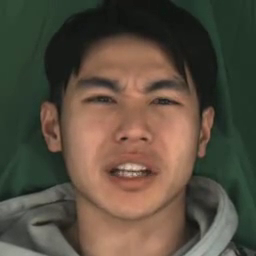}
    \end{minipage}
    \hspace{-1.3mm}
    \begin{minipage}[b]{0.13\textwidth}
        \centering
        \includegraphics[width=\textwidth]{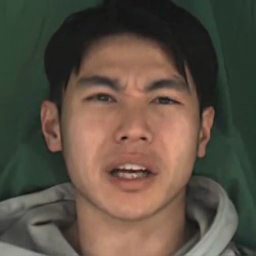}
    \end{minipage}
    \hspace{-1.3mm}
    \begin{minipage}[b]{0.13\textwidth}
        \centering
        \includegraphics[width=\textwidth]{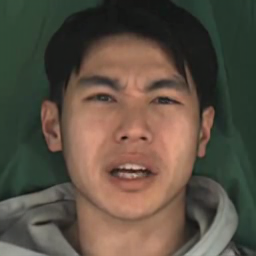}
    \end{minipage}
    \hspace{-1.3mm}
    \begin{minipage}[b]{0.13\textwidth}
        \centering
        \includegraphics[width=\textwidth]{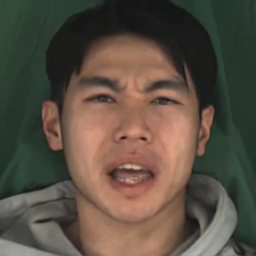}
    \end{minipage}

    \begin{minipage}[b]{0.13\textwidth}
        \centering
        \includegraphics[width=\textwidth]{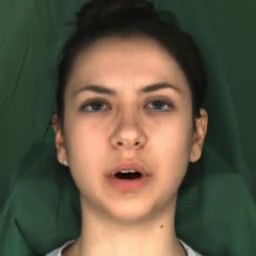}
        \subcaption*{Source}
    \end{minipage}
    \hspace{-1.3mm}
    \begin{minipage}[b]{0.13\textwidth}
        \centering
        \includegraphics[width=\textwidth]{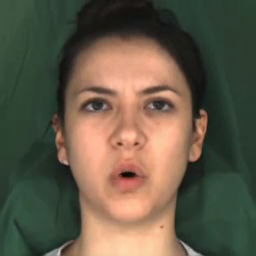}
        \subcaption*{5 frame}
    \end{minipage}
    \hspace{-1.3mm}
    \begin{minipage}[b]{0.13\textwidth}
        \centering
        \includegraphics[width=\textwidth]{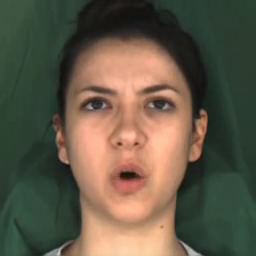}
        \subcaption*{10 frame}
    \end{minipage}
    \hspace{-1.3mm}
    \begin{minipage}[b]{0.13\textwidth}
        \centering
        \includegraphics[width=\textwidth]{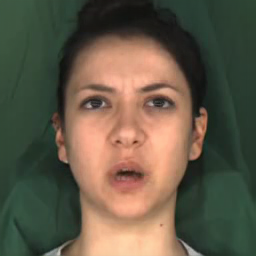}
        \subcaption*{50 frame}

    \end{minipage}
\caption{Several examples of predicting different frame numbers. The more frames predicted, the lower the image authenticity and expression similarity. }
\label{figure 1}
\end{figure}

On this basis, we propose an integrated framework in which SPFEM generates images with target expressions, and together with the corresponding audio segments, these are passed into an AD-THG model to yield precise lip synchronization. Nevertheless, our experiments reveal a critical limitation: as the sequence length increases, the quality and expression fidelity of AD-THG outputs tend to deteriorate, as shown in Figure \ref{figure 1}. To address this issue, we introduce an adjacent-frame learning strategy. Specifically, during training, the AD-THG model is optimized to predict $n$ consecutive neighboring frames, leveraging the temporal priors encoded in adjacent inputs. During inference, a frame produced by SPFEM and its corresponding $n$ frames of audio are provided as input, enabling the AD-THG model to generate $n$ consecutive high-quality frames. 
Unlike conventional spatio-temporal models that merely treat adjacent frames as additional temporal inputs, our method introduces an explicit adjacent-frame constraint that compels the network to learn frame-to-frame dependencies in a more structured and supervised fashion, thereby ensuring more stable, coherent, and high-quality generation results.

In summary, the main contributions of this work can be summarized as follows. 1) We propose a novel framework named Talking Head Facial Expression Manipulation (THFEM) that integrates SPFEM models with AD-THG models to better preserve the original mouth shape while manipulating facial expressions. 2) We propose an adjacent frame learning strategy, which leverages rich prior information of adjacent frames to further improve the quality of generations. 3) We conduct extensive experiments using various combinations of SPFEM and AD-THG models, demonstrating the effectiveness of our proposed framework. 
\emph{Code is publicly available at \href{https://github.com/liluoqaq/THFEM}{https://github.com/liluoqaq/THFEM}.} 

\section{Related Works}
\subsection{Talking Head Generation}
In recent years, research on deep learning in the field of talking head generation has attracted widespread attention. Here, we elaborate on two most closely related branches within the talking head generation domain: audio-driven talking head generation and expressive talking head generation. 

\noindent\textbf{Audio-driven Talking Head Generation.}
The audio-driven talking head task \cite{13,14,38,39,40,49,51}  aims to generate talking face videos from audio clips. Chen et al. \cite{11} designed a two-stage framework that leverages facial landmarks as intermediary representations to facilitate video generation. Zhou et al. \cite{9} decoupled speech content and speaker attributes from audio to produce speaker-aware talking head animations. Zhou et al. \cite{12} achieved free control of head pose when driving arbitrary talking faces with audios. Furthermore, Wang et al. \cite{10} proposed an audio-visual correlation transformer aimed at enhancing the realism of generated talking head videos. Despite these advances, control over facial expression remains limited. While initial methodologies emphasized lip synchronization, recent work has focused mainly on the generation of expressive talking head.

\noindent\textbf{Expressive Talking Head Generation.}
In the realm of animated presentations, realistic expressions are essential for conveying authenticity. Nevertheless, owing to the complexities associated with generating emotional information, previous research has rarely considered the integration of expressions in talking head generation. To bridge this gap, Sadoughi et al. \cite{15} adopted a conditional generative adversarial network based on the long short-term memory network to capture the intricate relationship between facial expression and mouth movement. On the other hand, Vougioukas et al. \cite{16} enhanced the realism of generated images by incorporating three discriminators into a temporal GAN framework. However, these methods still face obstacles in clearly conveying emotional semantics and facilitating expression manipulation. More recently, Ji et al. \cite{37} proposed the cross-reconstructed emotion disentanglement technique to decompose speech into two decoupled spaces and synthesize high-quality video portraits with vivid emotional dynamics driven by audio. Liang et al. \cite{18} trained an emotion and pose controllable model with a special pre-processing design. Ji et al. \cite{17} devised an implicit emotion disentanglement learning module to produce talking heads with controlled emotional expressions. Furthermore, Gan et al. \cite{19} incorporated an adaptive emotional module and a pre-trained emotion-agnostic talking head transformer to facilitate accurate emotional manipulation. 

\subsection{Facial Emotion Manipulation}
In this section, we explore two critical domains pertinent to the task of SPFEM: video-based face manipulation, and speech-preserving facial expression manipulation.

\noindent\textbf{Video-based Face Manipulation.}
In recent years, there has been a growing scholarly focus on video-based facial manipulation \cite{41,42,43,46}. Predominant studies in this area have employed conditional generative adversarial networks \cite{20,21,44,7} or 3D facial models (such as 3DMM) \cite{23} to modify facial attributes (e.g., hair color, gender) of speaking faces. These methodologies have facilitated visual transformations across various domains while preserving the essence of the original footage, marking a substantial contribution to the emerging field of deepfake technology. For instance, GANimation \cite{24} introduced a conditional GAN approach using Action Units \cite{25} to map facial movements that translate human expressions to a continuous domain. Furthermore, Tzaban et al. \cite{26} effectively harnessed the natural alignment characteristics of StyleGAN \cite{27} and the neural network's propensity for low-frequency functions to successfully learn the low-frequency representations, thereby achieving temporal consistency. Compared to traditional facial modification tasks, SPFEM presents a more formidable challenge since it requires modifying the facial expression while preserving the original speech animations.

\noindent\textbf{Speech Preserving Facial Expression Manipulation.}
The primary objective of SPFEM is to accurately transfer the source video to the target emotions while maintaining the speech-related facial movements. In previous research on SPFEM tasks such as the ICface \cite{1} project, expression editing was achieved through interpretable control signals like action units. However, this approach often disrupts the mouth shape when modifying expression, resulting in the inability to accurately preserve the original speech content. To overcome this challenge, Sun et al. \cite{28} employed 3DMM to decompose motion information into primary facial movements and utilized StyleGAN to refine texture maps that capture detailed facial appearance. Papantoniou et al.\cite{4} suggested a methodology for merging 3DMM parameters of both the source identity and target emotion to enhance expression fidelity. In our work, we introduce the audio-driven talking head model into the SPFEM task and propose an adjacent frame learning strategy to achieve better speech preservation during precise facial expression manipulation. 

\section{Method}
\begin{figure}[t]
  \centering
  \includegraphics[width=\textwidth]{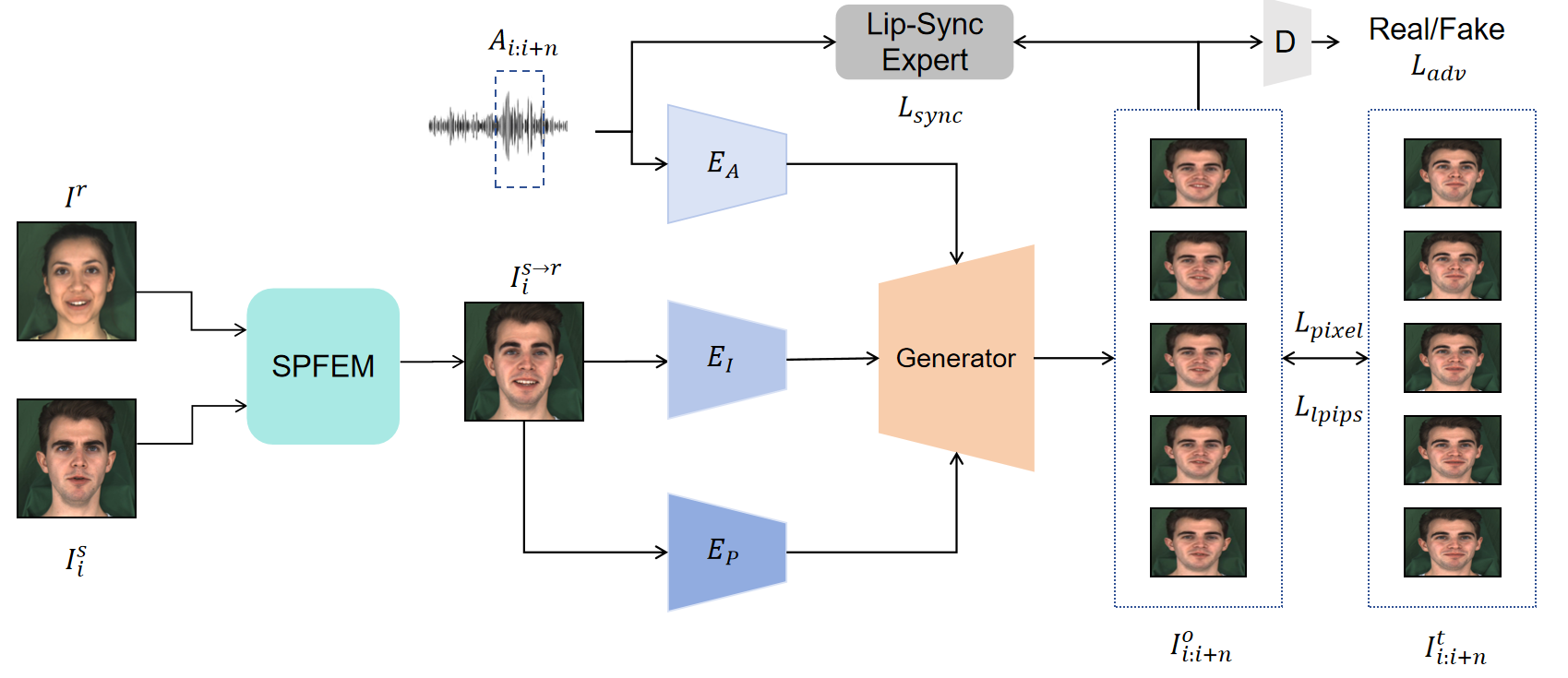} 
  \caption{An overall pipeline of the proposed THFEM that integrates SPFEM models with AD-THG models. Given a source image and a reference image, the SPFEM model first edits the source image to generate a facial image with the expression of the reference image. Subsequently, leveraging the output from SPFEM, the image encoder extracts identity information and pose encoder ascertains the head pose sequence, while the audio encoder derives mouth shape representations from the adjacent n frames of audio corresponding to the source image. Finally, the generator combines these three representations to predict n frame image sequence. (n is set to 5 in our experiments.)}
  \label{figure 2}
\end{figure}

We propose a novel framework called THFEM to improve the potential issue of mouth distortion during expression manipulation in the SPFEM task. As shown in Figure \ref{figure 2}, it encompasses two primary components: Firstly, it combines the AD-THG model and SPFEM model to preserve the natural contour of the original mouth by capitalizing on the strength of AD-THG in lip generation. Secondly, we implement an adjacent frame learning strategy that equips the AD-THG models with precise and comprehensive prior information. This enhancement allows the AD-THG models to more accurately predict subtle variations between neighboring frames, thus significantly improving the overall quality of the generated images.

\subsection{Preliminary}
In this section, we introduce the basic processes of the expression manipulation models and the audio-driven talking head models, which constitute the core components of our framework. 

\noindent\textbf{Facial Expression Manipulation Models.} Facial Expression Manipulation (FEM) models are specifically designed for video-to-video expression transfer. These models capture dynamic facial variations from a reference video and employ them to drive the source face, thereby facilitating precise and natural modification of facial expressions. For illustration, when given a single source image and a reference image , the expression manipulator predicts the target emotional representation based on the emotional information from both. Subsequently, the renderer leverages the original identity information and the synthesized expression representation to generate the facial image that contains the target emotion while preserving the original identity information. The process can be formulated as follows: 
\begin{equation}
\varepsilon = F(I^{s},I^{r})
\end{equation}
\begin{equation}
I^{s\rightarrow{r}} = R(I^{s},\varepsilon)
\end{equation}
where $I^{s}$, $I^{r}$ are respectively the source image and reference image, $F$ represents the expression manipulator and $R$ denotes the renderer.

\noindent\textbf{Audio-driven Talking Head Models.} Audio-driven talking head models aim to animate static images by accurately synchronizing lip movements with corresponding audio input. Specifically, this technology initiates by processing the audio through an audio encoder to extract audio embedding. Subsequently, the generator employs both the original image and extracted audio representation to synthesize lip movement animations that are aligned with the spoken content. This methodology can be formally expressed as follows:
\begin{equation}
I^{o} = G(I^{s}, E(A))
\end{equation}
where $I^{s}$, $A$ are respectively the source image and input audio, $E$ represents the audio encoder and $G$ denotes the generator.

\subsection{Efficient Integration of AD-THG and FEM models}
In the previous section, we introduced two models in detail, each with its unique advantages and limitations. Specifically, due to the complex coupling relationship between expressions and mouth shapes, as well as the lack of additional supervision for specifically mouth shape, existing FEM models excel in generating vivid and realistic videos but fall short in preserving detailed lip movements. Conversely, considering the high correlation between audio and mouth movement, the AD-THG models effectively generate lip movements that closely align with spoken content but tend to lack vividness in facial expressions. Considering the specific attributes of the FEM models and AD-THG models, we introduce audio information into the SPFEM task through the AD-THG models, enabling better preservation of mouth shapes that match the original speech content while editing facial expression. The pipeline of our framework is depicted in Figure \ref{figure 2}.

More concretely, our framework adopts a two-stage approach that efficiently integrates the functionalities of two distinct models: expression manipulation and audio-driven generation. Initially, given a source image $I^{s}$ and a reference image $I^{r}$, we employ Facial Expression Manipulation models $FEM$ to modify the source image, ensuring that the expression of the generated image aligns accurately with that of the reference image. In the second stage, we feed the image generated by the FEM model and the corresponding audio of source image into the AD-THG model $THG$ to facilitate more precise prediction of the subsequent sequence of images. Owing to the guidance of audio information and the robust lip-synthesis capabilities of the AD-THG model, the lip movements in the generated images are more accurately synchronized with the spoken content. The workflow of these two stages is depicted in the subsequent equation:
\begin{equation}
I^{s\rightarrow{r}} = FEM(I^{s},I^{r})
\end{equation}
\begin{equation}
I^{o} = THG(I^{s\rightarrow{r}}, E_{a}(A), E_{p}(I^{s\rightarrow{r}}))
\end{equation}
where $A$ is the input audio clip, and $E_{a}$, $E_{p}$ represent the audio encoder and head pose encoder, respectively. 

\subsection{Adjacent frame learning strategy}
After establishing the foundational framework, we utilize it to generate mouth shapes that are more accurately synchronized with the audio content. However, we observe that when the advanced AD-THG models synthesize an excessive number of frames simultaneously, there is a significant decline in both image quality and expression similarity. To mitigate this issue, we conduct an in-depth analysis of the AD-THG models and find that superior results are achieved when a reference image with highly relevant audio is used as input. To further enhance the realism and expression similarity of generated images, we introduce an advanced adjacent frame learning strategy to train the AD-THG models. 

The traditional AD-THG model is typically trained on real datasets. However, in our approach, we utilize the generated images by the FEM model as input for the AD-THG model. Therefore, it becomes imperative to leverage the FEM model to generate corresponding training data. We carefully select neutral videos as source videos and then utilize seven different emotional videos with the same ID as the drivers to generate the training dataset. To enhance the capacity of the  AD-THG model to precisely capture the relationship between audio and lip movements, we employ paired data (Videos with the same speak content but different emotion) for supervision. Specifically, we extract the mel-spectrograms \cite{47} of the generated video and its paired video. Subsequently, the dynamic programming algorithm \cite{48} is employed to perform precise frame alignment between these two videos along the time dimension on the spectrograms. This process ensures strict frame-to-frame correspondence. Based on the index of the source frame $I^{s}_{i}$, we can identify the target frame $I^{t}_{i}$ in the paired video that serves as supervision. This target frame mirrors the mouth shape of the original frame while precisely reflecting the intended reference emotion.

After successfully constructing the training dataset, we implement an advance adjacent frame learning strategy to train the AD-THG model. For each video, we encode its audio into a corresponding representation using an audio encoder. Subsequently, we deploy a head pose detector to ascertain the head pose sequence for each frame, a crucial element that is challenging to derive from audio alone but is vital for the authenticity of the video. During the training phase, for the i-th frame image of a given video, we consider it as the original image, providing essential identity information. Simultaneously, the corresponding audio and head pose sequences of the original image's adjacent n frames (i.e., from frame i+1 to i+n) are fed into the AD-THG model to make $n$ frame prediction. The parameter $n$ defines the temporal window of adjacent-frame learning, thereby controlling the amount of contextual information that can be effectively utilized. To  evaluate the impact of different $n$, we conduct ablation experiments under multiple settings, as shown in Table \ref{Table 5}. The results indicate that $n$ = 5 yields the most balanced trade-off between preserving perceptual quality and ensuring accurate lip-audio synchronization. Accordingly, we set $n$ = 5 in our experiments.

The training objective is constructed as a weighted sum of four complementary loss functions: synchronization loss, pixel reconstruction loss, perceptual loss, and adversarial loss. In accordance with standard practices in talking head generation, the weights of these losses are empirically adjusted to ensure balanced contributions during the optimization process. By minimizing this objective through backpropagation, our method achieves notable improvements in both visual fidelity and more accurate lip synchronization. In our experiments, the total loss function is defined as:
\begin{equation}
\mathcal{L} = \lambda_{sync} \mathcal{L}_{sync} + \lambda_{pixel}\mathcal{L}_{pixel} + \lambda_{lpips}\mathcal{L}_{lpips}+\lambda_{adv}\mathcal{L}_{dav}
\label{eq:loss-function}
\end{equation}
where $\lambda_{sync}$ , $\lambda_{pixel}$ , $\lambda_{lpips}$ and $\lambda_{adv}$ are balance coefficients that are set to 0.5, 10, 1, 0.1 in our experiments, respectively.

\noindent\textbf{Sync Loss}: The concept of synchronization loss is described in Wav2Lip \cite{29}. Based on the architecture of SyncNet \cite{30}, we leverage a pre-trained lip discriminator expert to assess lip synchronization across a variety of speaking styles. This specialized loss function enables the provision of precise guidance for lip movements, thereby facilitating an enhanced balance between style expressiveness and synchronization accuracy. For each batch of sampled audio clips, synchronization loss for the generated videos is computed using the following formula:
\begin{equation}
\mathcal{L}_{sync} = -\log(\frac{v \cdot s}{max(\|v\|_{2} \cdot \|s\|_{2}), \epsilon})
\end{equation}
The speech embedding $s$ and the generated video embedding $v$ are respectively extracted by the audio encoder and image encoder within the SyncNet framework. 

\noindent\textbf{Pixel Loss}: To enhance the quality of expression generation, we utilize an L1 reconstruction loss to provide pixel-level supervision, formulated as:
\begin{equation}
\mathcal{L}_{pixel} = \|I^{o} - I^{t} \|
\end{equation}

\noindent\textbf{Perceptual Loss}: Furthermore, to enhance the sharpness of the frames, we implement perceptual loss \cite{31} to the entire frames, denoted as:
\begin{equation}
\mathcal{L}_{lpips} = \|\phi_{p}(I^{o}) - \phi_{p}(I^{t}) \|
\end{equation}
where \text{$\phi_{p}$} denotes VGG19 \cite{32} for calculating perceptual loss.

\noindent\textbf{Generative Adversarial Loss}: We employ generative adversarial loss to ensure that the generated images more closely approximate the real data domain, thereby improving the realism of the images, formulated as: 

\begin{equation}
\mathcal{L}_{adv} = \min\limits_{G} \max\limits_{D} \mathbb{E}[\log(D(I^{s}))] + \mathbb{E}[\log(1 - D(I^{o}))]
\end{equation}

\section{Experiments}
\subsection{Experimental Settings}
\subsubsection{Dataset}
\
\newline
We conduct experiments on the MEAD \cite{33} dataset, which includes video clips of various emotions (i.e., neutral, angry, disgusted, fear, happy, sad, and surprised) from 60 actors. For the SPFEM model, we select videos of 6 speakers (M003, M009, W029, M012, M030, and W015) in various emotions, totaling 1,260 video samples. These were randomly divided, allocating 90\% for the training set and the remaining 10\% for the test set. For the AD-THG model, we utilize the data synthesized by the SPFEM model as input and employ real images for the supervision to facilitate effective training of the model.

\subsubsection{Evaluation Protocol}
\
\newline
In this work, we use the following metrics for evaluation. 1) Frechet Arcface Distance (FAD) \cite{34} employs the most advanced face recognition model \cite{35} to extract features from both generated and real videos. It then calculates the frechet distance between these features to assess the authenticity of the images. Lower FAD values indicate superior image quality. This metric evaluates not only the authenticity and clarity of images, but also the naturalness of generated images and the consistency of expression styles with real images. 2) Cosine Similarity (CSIM) \cite{36} gauges the facial expression cosine similarity between the generated video and the target emotional video using the latest facial expression recognition network. A higher CSIM value signifies greater similarity in facial expressions. 3) Lip Sync Error-Distance (LSE-D) \cite{29} utilizes a pre-trained model to accurately compute the distance between lip movements and audio representations, serving as an indicator of lip-audio synchronization. A smaller LSE-D corresponds to a higher lip-audio accuracy. We present the results of two experimental settings in detail. In the intra-identity setting, the source videos and expression references are from the same speaker, while in the cross-identity setting, they belong to different speakers. 

\subsection{Comparison with the Baseline Methods}
To validate the effectiveness of our proposed framework, we select two advanced SPFEM models as baselines. 1) DSM \cite{45} (ECCV 2020) performs semantic editing of facial videos based on categorical emotion labels, learning person-specific emotional representations in the Valence-Arousal space and rendering them as facial images. 2) NED (CVPR 2022) employs 3DMM parameters to establish the model for facial representation and achieves facial expression manipulation by modifying corresponding expression parameters. Meanwhile, we introduce the adjacent frame learning strategy for two state-of-the-art AD-THG models. 1) EAMM \cite{6} (SIGGRAPH 2022) proposes an Audio-to-Facial movement module to generate  neutral audio-driven talking faces by predicting unsupervised motion representations. 2) EAT (ICCV 2023) leverages the Audio-to-Expression Transformer to map audio to 3D key points and incorporates learnable guidance to steer facial generation.

\subsubsection{Quantitative Comparisons}
\ 
\newline
The results of the performance comparison are detailed in Table \ref{Table 1} and Table \ref{Table 2}. Both DSM and NED utilize 3DMM parameters to effectively edit facial expressions while preserving mouth movements. However, the intricate correlation between facial expression and mouth shape makes it impossible for limited-dimensional 3D parameters to fully decouple them, which results in sub-optimal image quality and lip preservation. In contrast, our framework leverages the AD-THG model's excellent lip generation capabilities and rich prior information from adjacent frames to generate more realistic images while better preserving the original lip movements.  
\begin{table}[h]
    \scalebox{0.7}{
    \setlength{\tabcolsep}{1pt}
    \begin{tabular}{ 
    c|ccc|ccc|ccc|ccc|ccc|ccc}
    \toprule
    \multirow{2}{*}{Emotions}  &\multicolumn{3}{c|}{DSM}  &\multicolumn{3}{c|}{DSM+EAMM}   &\multicolumn{3}{c|}{DSM+EAT}  &\multicolumn{3}{c|}{NED}  &\multicolumn{3}{c|}{NED+EAMM}  &\multicolumn{3}{c}{NED+EAT}\\
    \cline{2-19}
            & FAD$\downarrow$  & LSE-D$\downarrow$  & CSIM$\uparrow$  & FAD$\downarrow$    & LSE-D$\downarrow$  & CSIM$\uparrow$ & FAD$\downarrow$    & LSE-D$\downarrow$  & CSIM$\uparrow$ & FAD$\downarrow$    & LSE-D$\downarrow$  & CSIM$\uparrow$ & FAD$\downarrow$    & LSE-D$\downarrow$  & CSIM$\uparrow$ & FAD$\downarrow$    & LSE-D$\downarrow$  & CSIM$\uparrow$ \\
    \hline
     Neutral   & 0.572 & 9.692 & 0.907 & 1.628    & 9.446 & 0.869 & 0.857   & 9.380 & 0.911   & 0.657 & 9.764 & 0.918  & 1.518 & 9.543 & 0.881 & 1.028 & 9.437 & 0.908\\
    Angry     & 3.732 & 9.775 & 0.767 & 3.183    & 9.679 & 0.788 & 3.742   & 9.189 & 0.824   & 3.559 & 9.740 & 0.810  & 3.084 & 9.650 & 0.843 & 3.963 & 9.232 & 0.837\\
    Disgusted & 4.889 & 9.759 & 0.803 & 4.158    & 9.609 & 0.849 & 5.163   & 9.232 & 0.847   & 4.120 & 9.729 & 0.847  & 3.651 & 9.699 & 0.879 & 4.113 & 9.381 & 0.870\\
    Fear      & 4.539 & 9.731 & 0.757 & 3.777    & 9.541 & 0.788 & 3.315   & 9.187 & 0.859   & 4.495 & 9.724 & 0.790  & 3.653 & 9.475 & 0.823 & 3.136 & 9.191 & 0.870\\
    Happy     & 2.252 & 9.885 & 0.885 & 2.658    & 9.945 & 0.876 & 3.081   & 9.655 & 0.896   & 2.256 & 9.869 & 0.907  & 2.732 & 9.902 & 0.906 & 3.005 & 9.624 & 0.903\\
    Sad       & 4.432 & 9.843 & 0.754 & 4.494    & 9.731 & 0.788 & 3.754   & 9.616 & 0.860   & 4.267 & 9.864 & 0.766  & 4.115 & 9.526 & 0.822 & 3.653 & 9.673 & 0.849\\
    Surprised & 4.196 & 9.644 & 0.804 & 3.821    & 9.488 & 0.844 & 4.690   & 9.535 & 0.805   & 4.069 & 9.687 & 0.814  & 3.004 & 9.516 & 0.864 & 3.272 & 9.364 & 0.887\\
    Avg.      & 3.516 & 9.761 & 0.811 & 3.388    & 9.634 & 0.829 & 3.515   & 9.399 & 0.857   & 3.346 & 9.768 & 0.834  & 3.251 & 9.653 & 0.860 & 3.167 & 9.415 & 0.874\\

    \bottomrule
    \end{tabular}}
    \caption{Comparison results of  FAD, CSIM, and LES-D of NED, DSM with and without Neighboring AD-THG models in the intra-identity setting. }
    \label{Table 1}
\end{table}

\begin{table}[h]
    \scalebox{0.7}{
    \setlength{\tabcolsep}{1pt}
    \begin{tabular}{ 
    c|ccc|ccc|ccc|ccc|ccc|ccc}
    \toprule
    \multirow{2}{*}{Emotions}  &\multicolumn{3}{c|}{DSM}  &\multicolumn{3}{c|}{DSM+EAMM}   &\multicolumn{3}{c|}{DSM+EAT}  &\multicolumn{3}{c|}{NED}  &\multicolumn{3}{c|}{NED+EAMM}  &\multicolumn{3}{c}{NED+EAT}\\

    \cline{2-19}
    & FAD$\downarrow$  & LSE-D$\downarrow$  & CSIM$\uparrow$  & FAD$\downarrow$    & LSE-D$\downarrow$  & CSIM$\uparrow$ & FAD$\downarrow$    & LSE-D$\downarrow$  & CSIM$\uparrow$ & FAD$\downarrow$    & LSE-D$\downarrow$  & CSIM$\uparrow$ & FAD$\downarrow$    & LSE-D$\downarrow$  & CSIM$\uparrow$ & FAD$\downarrow$    & LSE-D$\downarrow$  & CSIM$\uparrow$ \\
    \hline
    Neutral   & 0.859 & 9.865 & 0.881 & 1.978    & 9.592 & 0.839 & 1.216   & 9.428 & 0.907  & 2.023 & 9.736 & 0.841 & 3.378 & 9.584 & 0.795 & 2.143 & 9.488 & 0.871\\
    Angry     & 4.067 & 9.843 & 0.725 & 3.435    & 9.670 & 0.748 & 3.820   & 9.140 & 0.819  & 4.852 & 9.805 & 0.717 & 4.385 & 9.783 & 0.716 & 5.174 & 9.264 & 0.815 \\
    Disgusted & 5.018 & 9.832 & 0.784 & 4.255    & 9.667 & 0.834 & 5.217   & 9.278 & 0.842  & 5.094 & 9.817 & 0.791 & 4.386 & 9.695 & 0.844 & 4.667 & 9.410 & 0.865\\
    Fear      & 4.986 & 9.740 & 0.746 & 4.027    & 9.578 & 0.769 & 3.323   & 9.148 & 0.860  & 4.983 & 9.767 & 0.750 & 4.503 & 9.692 & 0.798 & 3.420 & 9.245 & 0.871\\
    Happy     & 2.479 & 9.857 & 0.875 & 2.813    & 9.917 & 0.872 & 3.153   & 9.662 & 0.896  & 3.920 & 9.844 & 0.842 & 3.939 & 9.945 & 0.843 & 4.157 & 9.603 & 0.894\\
    Sad       & 4.999 & 9.775 & 0.722 & 5.089    & 9.734 & 0.758 & 4.034   & 9.595 & 0.856  & 5.665 & 9.926 & 0.691 & 5.338 & 9.761 & 0.753 & 4.862 & 9.731 & 0.837\\
    Surprised & 4.300 & 9.718 & 0.798 & 3.821    & 9.514 & 0.828 & 4.590   & 9.538 & 0.806  & 4.601 & 9.667 & 0.780 & 4.320 & 9.774 & 0.829 & 3.738 & 9.349 & 0.874\\
    Avg.      & 3.815 & 9.804 & 0.790 & 3.631    & 9.667 & 0.807 & 3.622   & 9.398 & 0.855  & 4.448 & 9.794 & 0.773 & 4.321 & 9.748 & 0.797 & 4.023 & 9.441 & 0.861\\
    \bottomrule 
    \end{tabular}}
    \caption{Comparision results of average FAD, CSIM, and LES-D of NED, DSM with and without Neighboring AD-THG models in the cross-identity setting. }
    \label{Table 2}
\end{table}
\ 
\newline
\indent
We first present the performance of the combination of the SPFEM models and the Neighboring AD-THG models in Table \ref{Table 1} and Table \ref{Table 2}. When the neighboring EAT model is integrated into NED, various evaluation metrics of generated images are improved. In the cross-identity setting, which is more general and practical setting, FAD, CSIM, LSE-D have all demonstrated a certain degree of improvement compared to NED itself, with FAD reduced from 4.448 to 4.023, LSE-D from 9.794 to 9.441, and CSIM from 0.773 to 0.861. These improvements, particularly in FAD and CSIM, can be attributed to the adjacent frame learning strategy employed during the training of the AD-THG models, which provides the models with rich and accurate prior information. This strategy enables the AD-THG models to capture subtle changes in facial expressions between adjacent frames more effectively, thus producing more vivid and realistic images. In addition, the lower LSE-D further demonstrates the effectiveness of the AD-THG models in maintaining lip-audio synchronization. In the intra-identity setting, these indicators also show notable improvement. The average FAD and LSE-D are reduced to 3.316 and 9.415, while CSIM increases to 0.874. This evidence demonstrates the flexibility of the neighboring AD-THG models in adapting to both intra-identity and cross-identity settings, thereby exhibiting superior generalization performance. When incorporating the neighboring EAT model into the DSM, in the cross-identity setting, compared to the  baseline DSM, which obtains average FAD, CSIM, LSE-D of 3.815, 9.804, 0.790, our framework decreases the average FAD, LSE-D to 3.622, 9.398, and increases the CSIM to 0.855. In the intra-identity setting, our framework has also resulted in a significant improvement on FAD, CSIM and LSE-D of different expression manipulation as well as the average FAD, LSE-D, and CSIM. 

To fully validate the generalization of the adjacent frame learning strategy, we train another neighboring AD-THG model EAMM to combine with SPFEM models and present the performance comparisons. As shown in Table \ref{Table 1} and Table \ref{Table 2}, incorporating the neighboring EAMM into SPFEM models  yields significant improvements in expression manipulation across  both settings. When the neighboring EAMM model is integrated into DSM in the intra-identity setting, it decreases the average FAD, LSE-D by 0.128 and 0.118, and increases the average CSIM by 0.018. In the cross-identity setting, it decreases the average FAD, LSE-D by 0.184 and 0.137, and increases the average CSIM by 0.017. When the neighboring EAMM model is integrated into NED , it achieves a comparable enhancement in performance metrics. In the intra-identity setting, our framework decreases the average FAD, LSE-D by 0.095 and 0.115, respectively, and increases the average CSIM by 0.026. In the cross-identity setting, it reduces the average FAD by 0.127 and increases the average CSIM by 0.024, while the average LSE-D remains the same. In summary, the test evaluation results of combining the two adjacent AD-THG models with the two SPFEM models demonstrate the effectiveness and versatility of our framework.

\begin{figure}[h]
    \centering
    \begin{minipage}[b]{0.13\textwidth}
        \centering
        \includegraphics[width=\textwidth]{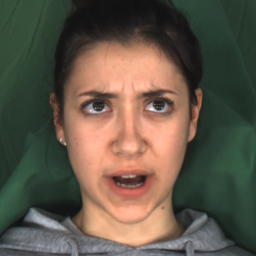}
    \end{minipage}
    \hspace{-1.3mm}
    \begin{minipage}[b]{0.13\textwidth}
        \centering
        \includegraphics[width=\textwidth]{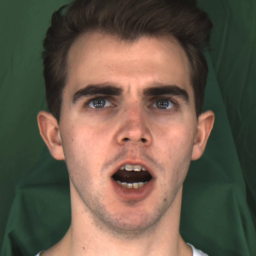}
    \end{minipage}
    \hspace{-1.3mm}
    \begin{minipage}[b]{0.13\textwidth}
        \centering
        \includegraphics[width=\textwidth]{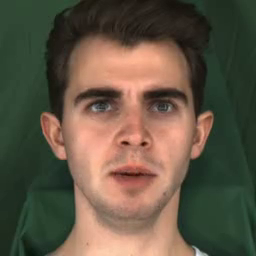}
    \end{minipage}
    \hspace{-1.3mm}
    \begin{minipage}[b]{0.13\textwidth}
        \centering
        \includegraphics[width=\textwidth]{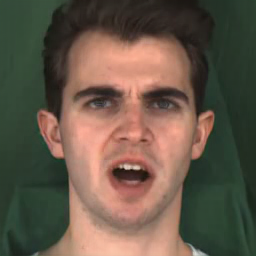}
    \end{minipage}
    \hspace{-1.3mm}
    \begin{minipage}[b]{0.13\textwidth}
        \centering
        \includegraphics[width=\textwidth]{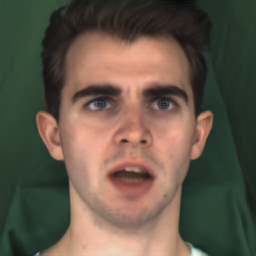}
    \end{minipage}
    
    \begin{minipage}[b]{0.13\textwidth}
        \centering
        \includegraphics[width=\textwidth]{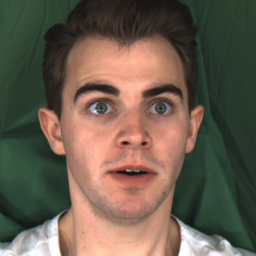}
    \end{minipage}
    \hspace{-1.3mm}
    \begin{minipage}[b]{0.13\textwidth}
        \centering
        \includegraphics[width=\textwidth]{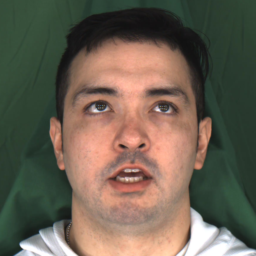}
    \end{minipage}
    \hspace{-1.3mm}
    \begin{minipage}[b]{0.13\textwidth}
        \centering
        \includegraphics[width=\textwidth]{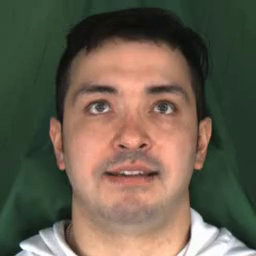}
    \end{minipage}
    \hspace{-1.3mm}
    \begin{minipage}[b]{0.13\textwidth}
        \centering
        \includegraphics[width=\textwidth]{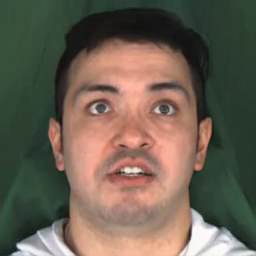}
    \end{minipage}
    \hspace{-1.3mm}
    \begin{minipage}[b]{0.13\textwidth}
        \centering
        \includegraphics[width=\textwidth]{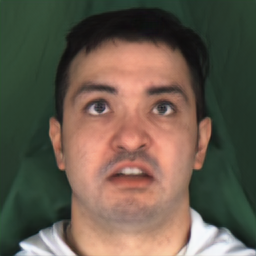}
    \end{minipage}

    \begin{minipage}[b]{0.13\textwidth}
        \centering
        \includegraphics[width=\textwidth]{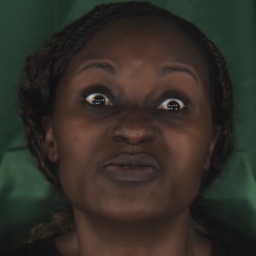}
    \end{minipage}
    \hspace{-1.3mm}
    \begin{minipage}[b]{0.13\textwidth}
        \centering
        \includegraphics[width=\textwidth]{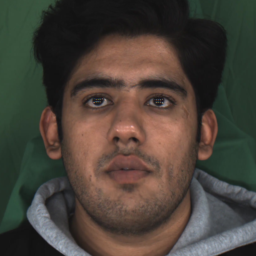}
    \end{minipage}
    \hspace{-1.3mm}
    \begin{minipage}[b]{0.13\textwidth}
        \centering
        \includegraphics[width=\textwidth]{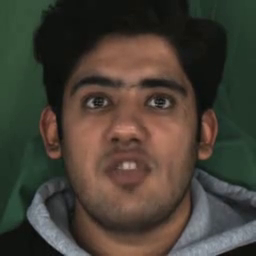}
    \end{minipage}
    \hspace{-1.3mm}
    \begin{minipage}[b]{0.13\textwidth}
        \centering
        \includegraphics[width=\textwidth]{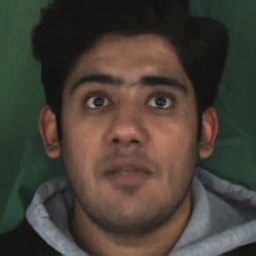}
    \end{minipage}
    \hspace{-1.3mm}
    \begin{minipage}[b]{0.13\textwidth}
        \centering
        \includegraphics[width=\textwidth]{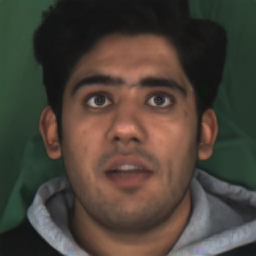}
    \end{minipage}

    \begin{minipage}[b]{0.13\textwidth}
        \centering
        \includegraphics[width=\textwidth]{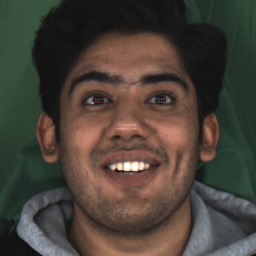}
    \end{minipage}
    \hspace{-1.3mm}
    \begin{minipage}[b]{0.13\textwidth}
        \centering
        \includegraphics[width=\textwidth]{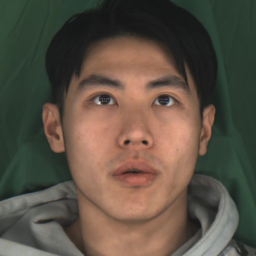}
    \end{minipage}
    \hspace{-1.3mm}
    \begin{minipage}[b]{0.13\textwidth}
        \centering
        \includegraphics[width=\textwidth]{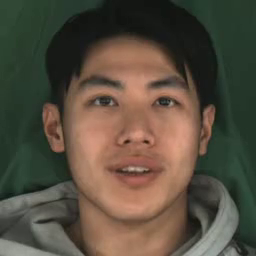}
    \end{minipage}
    \hspace{-1.3mm}
    \begin{minipage}[b]{0.13\textwidth}
        \centering
        \includegraphics[width=\textwidth]{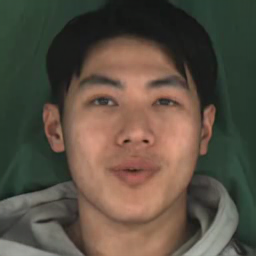}
    \end{minipage}
    \hspace{-1.3mm}
    \begin{minipage}[b]{0.13\textwidth}
        \centering
        \includegraphics[width=\textwidth]{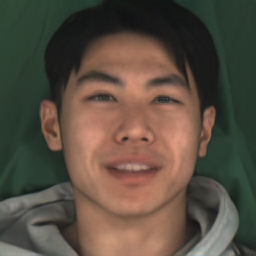}
    \end{minipage}

    \begin{minipage}[b]{0.13\textwidth}
        \centering
        \includegraphics[width=\textwidth]{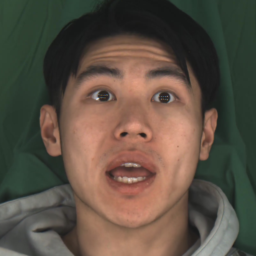}
    \end{minipage}
    \hspace{-1.3mm}
    \begin{minipage}[b]{0.13\textwidth}
        \centering
        \includegraphics[width=\textwidth]{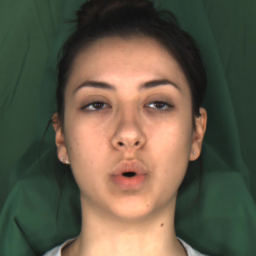}
    \end{minipage}
    \hspace{-1.3mm}
    \begin{minipage}[b]{0.13\textwidth}
        \centering
        \includegraphics[width=\textwidth]{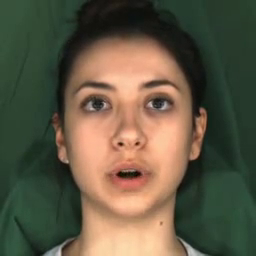}
    \end{minipage}
    \hspace{-1.3mm}
    \begin{minipage}[b]{0.13\textwidth}
        \centering
        \includegraphics[width=\textwidth]{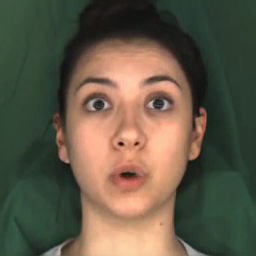}
    \end{minipage}
    \hspace{-1.3mm}
    \begin{minipage}[b]{0.13\textwidth}
        \centering
        \includegraphics[width=\textwidth]{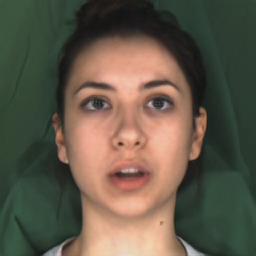}
    \end{minipage}

    \begin{minipage}[b]{0.13\textwidth}
        \centering
        \includegraphics[width=\textwidth]{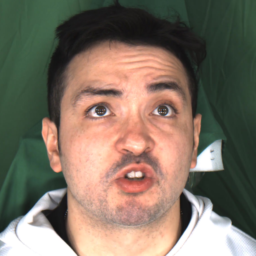}
        \subcaption*{Reference}
    \end{minipage}
    \hspace{-1.3mm}
    \begin{minipage}[b]{0.13\textwidth}
        \centering
        \includegraphics[width=\textwidth]{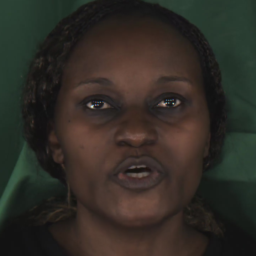}
        \subcaption*{Source}
    \end{minipage}
    \hspace{-1.3mm}
    \begin{minipage}[b]{0.13\textwidth}
        \centering
        \includegraphics[width=\textwidth]{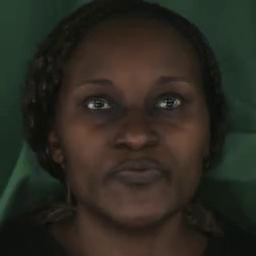}
        \subcaption*{NED}
    \end{minipage}
    \hspace{-1.3mm}
    \begin{minipage}[b]{0.13\textwidth}
        \centering
        \includegraphics[width=\textwidth]{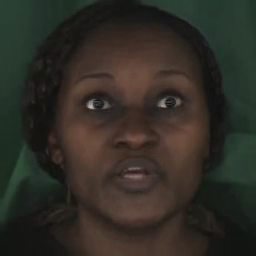}
        \subcaption*{NED+EAT}
    \end{minipage}
    \hspace{-1.3mm}
    \begin{minipage}[b]{0.13\textwidth}
        \centering
        \includegraphics[width=\textwidth]{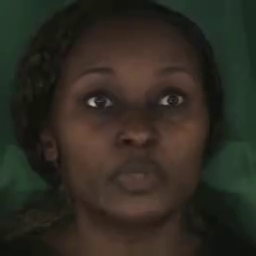}
        \subcaption*{NED+EAMM}
    \end{minipage}
\caption{Qualitative comparisons of NED with and without the neighboring AD-THG models.}
\label{figure 3}
\end{figure}

\begin{figure}[b]
    \centering
    \begin{minipage}[b]{0.13\textwidth}
        \centering
        \includegraphics[width=\textwidth]{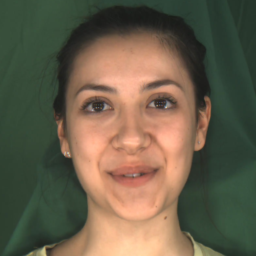}
    \end{minipage}
    \hspace{-1.3mm}
    \begin{minipage}[b]{0.13\textwidth}
        \centering
        \includegraphics[width=\textwidth]{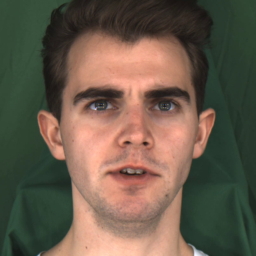}
    \end{minipage}
    \hspace{-1.3mm}
    \begin{minipage}[b]{0.13\textwidth}
        \centering
        \includegraphics[width=\textwidth]{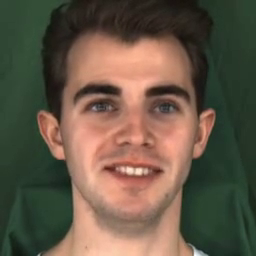}
    \end{minipage}
    \hspace{-1.3mm}
    \begin{minipage}[b]{0.13\textwidth}
        \centering
        \includegraphics[width=\textwidth]{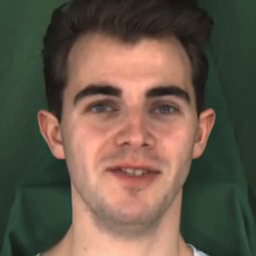}
    \end{minipage}
    \hspace{-1.3mm}
    \begin{minipage}[b]{0.13\textwidth}
        \centering
        \includegraphics[width=\textwidth]{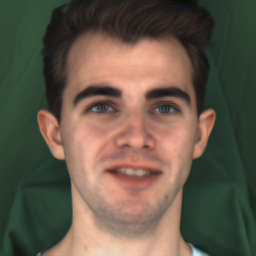}
    \end{minipage}

    \begin{minipage}[b]{0.13\textwidth}
        \centering
        \includegraphics[width=\textwidth]{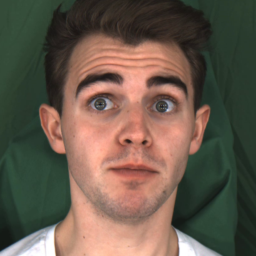}
    \end{minipage}
    \hspace{-1.3mm}
    \begin{minipage}[b]{0.13\textwidth}
        \centering
        \includegraphics[width=\textwidth]{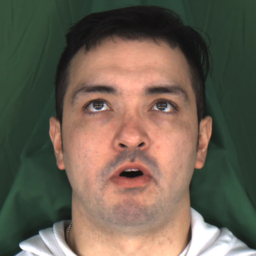}
    \end{minipage}
    \hspace{-1.3mm}
    \begin{minipage}[b]{0.13\textwidth}
        \centering
        \includegraphics[width=\textwidth]{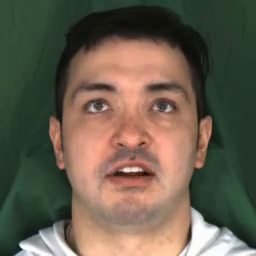}
    \end{minipage}
    \hspace{-1.3mm}
    \begin{minipage}[b]{0.13\textwidth}
        \centering
        \includegraphics[width=\textwidth]{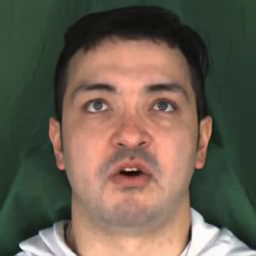}
    \end{minipage}
    \hspace{-1.3mm}
    \begin{minipage}[b]{0.13\textwidth}
        \centering
        \includegraphics[width=\textwidth]{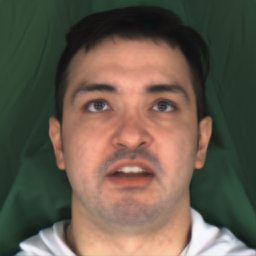}
    \end{minipage}

    \begin{minipage}[b]{0.13\textwidth}
        \centering
        \includegraphics[width=\textwidth]{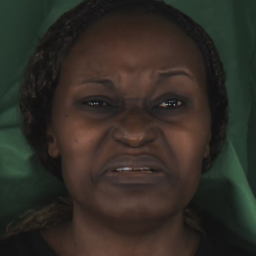}
    \end{minipage}
    \hspace{-1.3mm}
    \begin{minipage}[b]{0.13\textwidth}
        \centering
        \includegraphics[width=\textwidth]{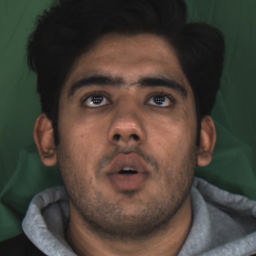}
    \end{minipage}
    \hspace{-1.3mm}
    \begin{minipage}[b]{0.13\textwidth}
        \centering
        \includegraphics[width=\textwidth]{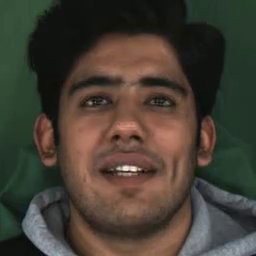}
    \end{minipage}
    \hspace{-1.3mm}
    \begin{minipage}[b]{0.13\textwidth}
        \centering
        \includegraphics[width=\textwidth]{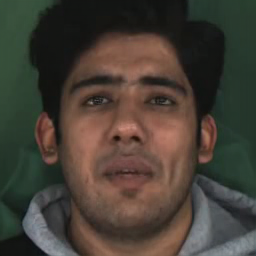}
    \end{minipage}
    \hspace{-1.3mm}
    \begin{minipage}[b]{0.13\textwidth}
        \centering
        \includegraphics[width=\textwidth]{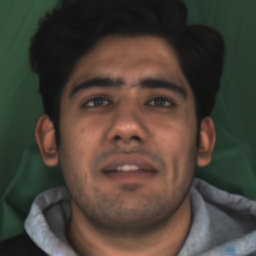}
    \end{minipage}

    \begin{minipage}[b]{0.13\textwidth}
        \centering
        \includegraphics[width=\textwidth]{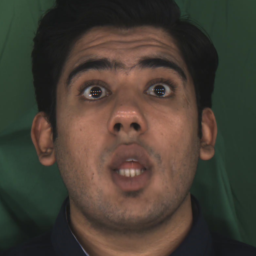}
    \end{minipage}
    \hspace{-1.3mm}
    \begin{minipage}[b]{0.13\textwidth}
        \centering
        \includegraphics[width=\textwidth]{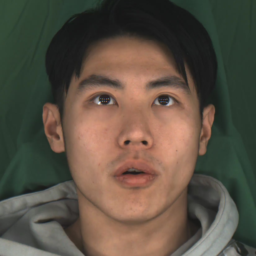}
    \end{minipage}
    \hspace{-1.3mm}
    \begin{minipage}[b]{0.13\textwidth}
        \centering
        \includegraphics[width=\textwidth]{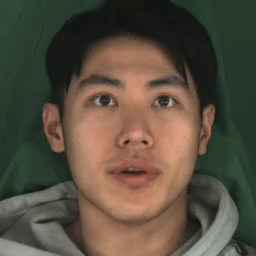}
    \end{minipage}
    \hspace{-1.3mm}
    \begin{minipage}[b]{0.13\textwidth}
        \centering
        \includegraphics[width=\textwidth]{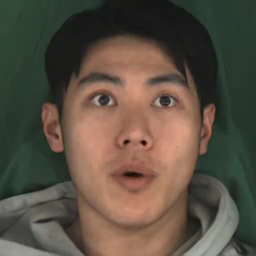}
    \end{minipage}
    \hspace{-1.3mm}
    \begin{minipage}[b]{0.13\textwidth}
        \centering
        \includegraphics[width=\textwidth]{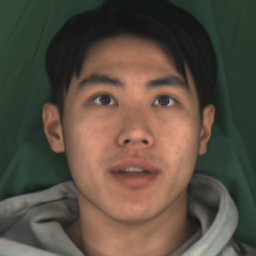}
    \end{minipage}

    \begin{minipage}[b]{0.13\textwidth}
        \centering
        \includegraphics[width=\textwidth]{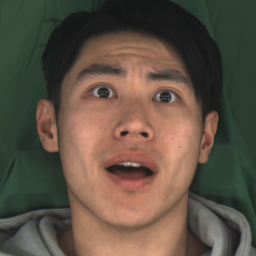}
    \end{minipage}
    \hspace{-1.3mm}
    \begin{minipage}[b]{0.13\textwidth}
        \centering
        \includegraphics[width=\textwidth]{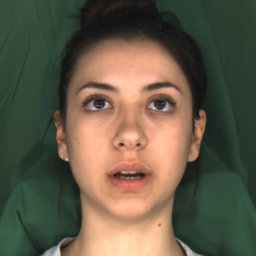}
    \end{minipage}
    \hspace{-1.3mm}
    \begin{minipage}[b]{0.13\textwidth}
        \centering
        \includegraphics[width=\textwidth]{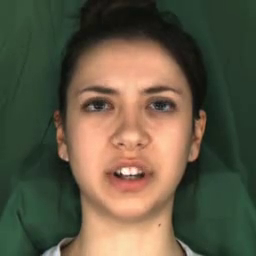}
    \end{minipage}
    \hspace{-1.3mm}
    \begin{minipage}[b]{0.13\textwidth}
        \centering
        \includegraphics[width=\textwidth]{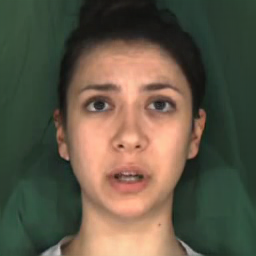}
    \end{minipage}
    \hspace{-1.3mm}
    \begin{minipage}[b]{0.13\textwidth}
        \centering
        \includegraphics[width=\textwidth]{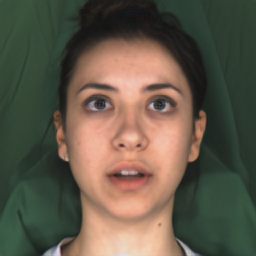}
    \end{minipage}

    \begin{minipage}[b]{0.13\textwidth}
        \centering
        \includegraphics[width=\textwidth]{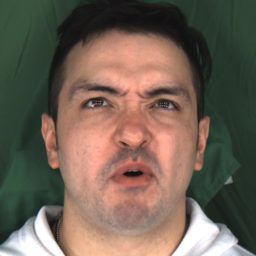}
        \subcaption*{Reference}
    \end{minipage}
    \hspace{-1.3mm}
    \begin{minipage}[b]{0.13\textwidth}
        \centering
        \includegraphics[width=\textwidth]{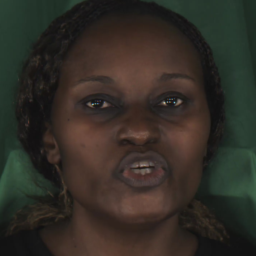}
        \subcaption*{Source}
    \end{minipage}
    \hspace{-1.3mm}
    \begin{minipage}[b]{0.13\textwidth}
        \centering
        \includegraphics[width=\textwidth]{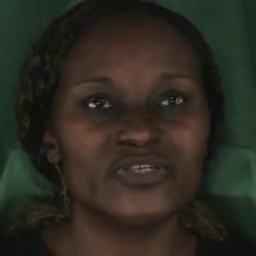}
        \subcaption*{DSM}
    \end{minipage}
    \hspace{-1.3mm}
    \begin{minipage}[b]{0.13\textwidth}
        \centering
        \includegraphics[width=\textwidth]{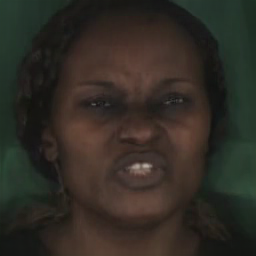}
        \subcaption*{DSM+EAT}
    \end{minipage}
    \hspace{-1.3mm}
    \begin{minipage}[b]{0.13\textwidth}
        \centering
        \includegraphics[width=\textwidth]{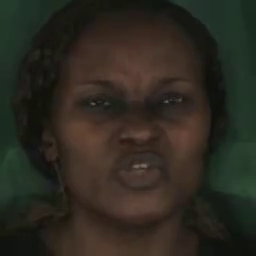}
        \subcaption*{DSM+EAMM}
    \end{minipage}
\caption{Qualitative comparisons of DSM with and without the neighboring AD-THG models. }
\label{figure 4}
\end{figure}

\subsubsection{Qualitative Comparisons}
\ 
\newline
In this section, we present some visualization results of the NED and DSM, both with and without the neighboring AD-THG models, as illustrated in Figure \ref{figure 3} and Figure \ref{figure 4}. During the qualitative comparison process, we employ a methodology analogous to that used in our quantitative analyses, focusing primarily on the following three dimensions. 1) \textbf{Realism}: The images generated by NED have unnatural mouth regions due to inaccurate lip shape prediction, as evidenced in the second and third rows of Figure \ref{figure 3}. The images produced  by DSM exhibit unnatural distortions in the eye and mouth regions, as evidenced  in the fifth row of Figure \ref{figure 4}. Contrarily, our framework generates the faces with higher fidelity. 2) \textbf{Lip-audio preserving accuracy}: Given the inherent difficulties in decoupling facial expression and mouth shape features, current methods often inevitably lead to distortions in mouth shape when editing expressions. This is evident in the examples in the first and sixth rows of Figure \ref{figure 3}, where the mouth shapes in images generated by NED differ significantly from the source images. In addition, in the first and third rows of Figure \ref{figure 4}, the mouth shape of the generated images is not well preserved. In contrast, we introduce neighboring AD-THG models which can accurately generate lip movements that correspond to the original audio. 3) \textbf{Emotion similarity}: The existing approaches utilize 3DMM parameters as emotional representations, which is insufficient to capture complex facial emotion. These methods often tend to replicate reference expression onto the source image, resulting in significant differences in facial expression compared to the reference image, as depicted in the second and fifth rows of Figure \ref{figure 3}. More seriously, the SPFEM models occasionally even generate images with expressions that are vastly different from the reference image, evident in the third and fourth rows of Figure \ref{figure 4}. Conversely, Our framework harnesses the rich prior information of adjacent frames, thereby considerably enhancing the precision of the generated expressions. \emph{For a more intuitive comparison, we have also included some video comparisons in \href{https://liluoqaq.github.io/THFEM-Supp/}{https://liluoqaq.github.io/THFEM-Supp/}.}

\begin{table}[h]
    \scalebox{0.9}{
    \setlength{\tabcolsep}{1pt}
    \begin{tabular}{ 
    c|c|ccc|ccc|ccc|ccc}
    \toprule
    \multirow{2}{*}{Setting}  &\multirow{2}{*}{Emotions} &\multicolumn{3}{c|}{SSERD}  &\multicolumn{3}{c|}{SSERD+EAT}   &\multicolumn{3}{c|}{CDRL}  &\multicolumn{3}{c}{CDRL+EAT}\\

    \cline{3-14}
    & 
    & FAD$\downarrow$  & LSE-D$\downarrow$  & CSIM$\uparrow$  & FAD$\downarrow$    & LSE-D$\downarrow$  & CSIM$\uparrow$ & FAD$\downarrow$    & LSE-D$\downarrow$  & CSIM$\uparrow$ & FAD$\downarrow$    & LSE-D$\downarrow$  & CSIM$\uparrow$ \\
    \hline
    \multirow{8}{*}{Intra-ID}
        &Neutral   & 0.620 & 9.414 & 0.921   & 1.241 & 9.383 & 0.918    & 0.866 & 9.265 & 0.929  & 1.288 & 9.277 & 0.920 \\
        &Angry     & 3.419 & 9.289 & 0.822   & 2.881 & 9.152 & 0.843    & 3.236 & 9.159 & 0.857  & 3.134 & 9.005 & 0.846 \\
        &Disgusted & 4.023 & 9.333 & 0.876   & 3.859 & 9.141 & 0.886    & 4.480 & 9.216 & 0.857  & 4.432 & 9.090 & 0.868 \\
        &Fear      & 2.481 & 9.496 & 0.876   & 2.512 & 9.266 & 0.899    & 4.571 & 9.286 & 0.814  & 3.898 & 9.195 & 0.868 \\
        &Happy     & 2.552 & 9.663 & 0.890   & 2.539 & 9.608 & 0.900    & 2.986 & 9.389 & 0.900  & 2.973 & 9.466 & 0.890 \\
        &Sad       & 2.742 & 9.487 & 0.875   & 2.661 & 9.430 & 0.881    & 4.225 & 9.421 & 0.837  & 3.707 & 9.348 & 0.853 \\
        &Surprised & 2.460 & 9.442 & 0.888   & 2.444 & 9.304 & 0.905    & 4.264 & 9.236 & 0.831  & 4.069 & 9.382 & 0.874 \\
        &Avg.      & 2.614 & 9.446 & 0.878   & 2.591 & 9.326 & 0.890    & 3.518 & 9.282 & 0.861  & 3.357 & 9.252 & 0.874 \\
    \cline{1-14}
    \multirow{8}{*}{Cross-ID}
        &Neutral   & 3.074 & 9.894 & 0.862  & 3.412  & 9.592 & 0.880    & 1.872 & 9.329 & 0.874  & 2.058 & 9.347 & 0.899 \\
        &Angry     & 3.669 & 9.842 & 0.794  & 3.544  & 9.670 & 0.817    & 4.862 & 9.319 & 0.781  & 4.381 & 9.059 & 0.815 \\
        &Disgusted & 4.420 & 9.762 & 0.824  & 4.331  & 9.667 & 0.851    & 4.933 & 9.265 & 0.845  & 4.741 & 9.112 & 0.870 \\
        &Fear      & 3.951 & 9.968 & 0.831  & 4.064  & 9.578 & 0.870    & 5.252 & 9.301 & 0.795  & 4.411 & 9.257 & 0.863 \\
        &Happy     & 4.091 & 9.907 & 0.838  & 3.906  & 9.917 & 0.875    & 3.584 & 9.333 & 0.880  & 3.279 & 9.282 & 0.889 \\
        &Sad       & 4.124 & 9.959 & 0.833  & 3.900  & 9.734 & 0.851    & 5.394 & 9.472 & 0.780  & 4.967 & 9.407 & 0.817 \\
        &Surprised & 3.906 & 9.923 & 0.835  & 4.034  & 9.514 & 0.888    & 4.768 & 9.252 & 0.811  & 4.393 & 9.286 & 0.870 \\
        &Avg.      & 3.891 & 9.894 & 0.831  & 3.884  & 9.667 & 0.862    & 4.381 & 9.324 & 0.824  & 4.033 & 9.250 & 0.860 \\
    \bottomrule 
    \end{tabular}
    \color{black}}
    \caption{Comparision results of average FAD, CSIM, and LES-D of SSERD, CDRL with and without Neighboring EAT model in the intra-identity and cross-identity setting. }
    \label{Table 3}
\end{table}

\subsection{Generalization to the Latest Methods}
To further evaluate the generalization capability of our framework, we select two state-of-the-art SPFEM models as baselines and incorporate EAT with the adjacent frame training strategy as the talking-head module. Specifically, 1) SSERD \cite{xu2024self} performs expression editing by manipulating the latent code and then employs StyleGAN to generate the final images; 2) CDRL \cite{chen2025contrastive} introduces two dedicated modules, namely Contrastive Content Representation Learning module and Contrastive Emotion Representation Learning module, to disentangle content and expression from reference images, thereby providing more effective guidance for SPFEM training.
\newline
\indent
We systematically evaluate SSERD and CDRL with and without the neighboring EAT module under both intra-identity and cross-identity conditions, as summarized in Table \ref{Table 3}. In the intra-identity setting, incorporating the neighboring EAT module consistently improves both baselines. Compared to SSERD alone, the FAD score decreases from 2.614 to 2.591, the LSE-D decreases from 9.446 to 9.326, and the CSIM increases from 0.878 to 0.890. For CDRL, the integration of the neighboring EAT module reduces FAD from 3.518 to 3.357, decreases LSE-D from 9.282 to 9.252, and raises CSIM from 0.861 to 0.874. In the more challenging cross-identity setting, the gains are even more pronounced, underscoring the generalization capability of our framework. Relative to SSERD without EAT, the FAD remains stable at approximately 3.89, while the LSE-D decreases from 9.894 to 9.667 and the CSIM increases from 0.831 to 0.862. Similarly, compared with CDRL alone, the integration of neighboring EAT reduces FAD from 4.381 to 4.033, decreases LSE-D from 9.324 to 9.250, and improves CSIM from 0.824 to 0.860. Taken together, these results confirm that our approach consistently enhances image fidelity, improves synchronization accuracy, and achieves higher expression similarity across both intra- and cross-identity scenarios. 
\newline
\indent
Overall, these results demonstrate the strong generalization of our framework when applied to recent state-of-the-art SPFEM models. By integrating the adjacent frame EAT module into these baselines, the system consistently achieves lower FAD, improved synchronization accuracy, and higher expression similarity. These consistent gains across both intra-identity and cross-identity settings highlight not only the robustness of the proposed approach but also its effectiveness in enhancing the performance of advanced SPFEM architectures. 

\subsection{User Study}
We conduct a web-based user study to evaluate the comparative performance of our framework against established baselines. It consists of three parts, each corresponding to one of the following metrics: realism, emotional similarity to the reference emotion, and lip-sync similarity to the source video. For each combination of models, we carefully select 14 videos, encompassing seven distinct emotions, thus obtaining 56 videos. We find 20 participants to assess the three dimensions of each video.  

As detailed in Table \ref{Table 4},  our framework consistently outperforms the baselines across all three evaluated metrics. Specifically, for each model combination within our framework, there are notable improvements over baseline performance in realism, emotional similarity, and mouth shape similarity. The combination of the neighboring EAMM model and DSM yields substantial enhancements, achieving ratings that are 30\% higher in realism, 22\% higher in emotional similarity, and 30\% higher in mouth shape similarity compared to the baseline. Similarly, when DSM and the neighboring EAT model are combined, the improvements are consistent: 18\% higher in realism, 20\% higher in emotional similarity, and 14\% higher in mouth shape similarity relative to the baseline.

On the other hand, when NED is integrated with the neighboring EAMM model, our framework achieves superior performance, with ratings of 60\% for realism, 62\% for emotion similarity, and 57\% for mouth shape similarity, outperforming NED by 40\%, 38\%, and 43\%, respectively. Additionally, when NED is combined with the neighboring EAT model, it demonstrates notable improvements, with a 14\% higher rating in realism, a 26\% higher in emotion similarity, and a 12\% higher rating in mouth shape similarity. These results decisively demonstrate that our methodology significantly enhances the performance of the SPFEM models.

\begin{table}[h]
    \setlength{\tabcolsep}{1.5pt}{
    \begin{tabular}{c|cc|cc|cc|cc}
    \toprule
    Metrics & DSM & DSM+EAMM & DSM & DSM+EAT & NED & NED+EAMM & NED & NED+EAT \\
    \hline

    Realism & 35$\%$ & 65$\%$ & 46$\%$ & 54$\%$ & 40$\%$ & 60$\%$ & 43$\%$ & 57$\%$ \\

    Emotion similarity & 39$\%$ & 61$\%$ & 40$\%$ & 60$\%$ & 38$\%$ & 62$\%$ & 37$\%$ & 63$\%$ \\

    Mouth shape similarity & 35$\%$ & 65$\%$ & 48$\%$ & 52$\%$ & 43$\%$ & 57$\%$ & 44$\%$ & 56$\%$ \\

    \bottomrule
    \end{tabular}}
    \caption{Realism, emotion similarity, and mouth shape similarity ratings of the user study.}
    \label{Table 4}
\end{table}

\subsection{Ablation Study}
In this section, we first conduct a comprehensive analysis of the performance of the latest AD-THG model in the context of emotional talking head generation. Following that, we delve into the specific impact of the number of frames predicted by the neighboring frame model on the quality of the generated images. These experiments are performed using the EAT model.
 
\subsubsection{Analysis of the integration of AD-THG models}
\
\newline
The EAT incorporates an adaptive emotional module into an emotion-agnostic talking head model to achieve expression manipulation. However, since the EAT model utilizes textual labels for emotional manipulation without incorporating identity information, we perform a comparative analysis with other methodologies under a cross-identity setting to ensure a comprehensive and equitable comparison.

We employ four methods to edit neutral images and analyze their editing effects from three dimensions: realism, lip synchronization, and expression similarity. As illustrated in Table \ref{Table 5}, compared to NED, EAT demonstrates significant advantages in lip synchronization; however, it falls short in maintaining the authenticity of the generated images. The SPFEM model and the AD-THG model each possess distinct strengths in lip generation and image manipulation, respectively. Our framework effectively combines the advantages of both, thereby enhancing the performance of the SPFEM model.

In addition, we evaluate the effectiveness of the adjacent frame learning strategy. Compared to the NED+EAT model trained without this strategy, our approach achieves notable performance gains. As shown in Table \ref{Table 5}, FAD decreases from 6.638 to  4.023 and CSIM increases from 0.738 to 0.861, indicating substantial improvements in visual quality. Although LSE-D exhibits a slight increase, the overall synchronization remains stable. These results demonstrate the pivotal role of the adjacent frame learning in enhancing both image quality and the consistency of facial expressions. 

\begin{table}[h]
\centering
\begin{tabular}{c|c|ccc}
\toprule
Setting                     & Methods & FAD↓ & LSE-D↓ & CSIM↑ \\
\hline

\multirow{4}{*}{Cross-ID}   & NED       & 4.448 & 9.794 & 0.773 \\
                            & EAT       & 5.085 & \textbf{9.093} & 0.796 \\
                            & NED+EAT(w/o adjacent frame prior)   & 6.638 & 9.216 & 0.783 \\
                            & NED+EAT(with adjacent frame prior)  & \textbf{4.023} & 9.441 & \textbf{0.861} \\
\bottomrule 
\end{tabular}
\caption{Comparison results of average FAD, LSE-D, and CSIM of NED, EAT and our framework in cross-identity  settings.}
\label{Table 5}
\end{table}

\subsubsection{Analysis of Number of predicted frames}
\ 
\newline
Neighboring frames can furnish the AD-THG model with rich prior information, markedly improving the quality of facial generation. In our research, we conduct an in-depth analysis of the impact of varying frame counts during the inference process.  According to Table \ref{Table 6}, increasing the number of frames generated concurrently tends to reduce the authenticity of the generated images and the similarity of expressions. Specifically, in both settings, when the number of inference frames increased from 5 to 50, we observe a significant decrease in the realism of generated images and expression similarity, alongside a slight improvement in lip synchronization. As shown in Table \ref{Table 6} with the intra-ID setting, compared to the fifty frame model, which obtains average FAD, LSE-D, and CSIM of 4.290, 9.355, and 0.842, the five frame model decreases the average FAD to 3.167, with a relative decrement of 26.2\%, and increases the average LSE-D and CSIM to 9.415 and 0.874, corresponding to relative increments of 0.64\% and 3.8\%. In the cross-ID setting, the changes in metrics are similar. In conclusion, despite a slight reduction in lip-audio synchronization accuracy, the enhancements in image realism and emotional similarity are significant, confirming that our current configuration is both reasonable and effective. 

\begin{table}[h]
\begin{tabular}{c|c|ccc}
\toprule
\multirow{2}{*}{Setting}  & Number of  & \multirow{2}{*}{FAD↓}  & \multirow{2}{*}{LSE-D↓} & \multirow{2}{*}{CSIM↑}  \\
 &predicted frames  &  &  & \\
\hline
\multirow{3}{*}{Intra-ID}   & 5    &\textbf{3.167} & 9.415 & \textbf{0.874} \\
                            & 20   & 3.408 & 9.392 & 0.865 \\
                            & 50   & 4.290 & \textbf{9.355} & 0.842 \\
\hline
\multirow{3}{*}{Cross-ID}   & 5    & \textbf{4.023} & 9.441 & \textbf{0.861} \\
                            & 20   & 4.209 & 9.397 & 0.854 \\
                            & 50   & 5.017 & \textbf{9.340} & 0.832 \\
\bottomrule 
\end{tabular}
\caption{Comparison results of average FAD, LSE-D, and CSIM for different numbers of predicted frames. }
\label{Table 6}
\end{table}

\section{Limitation}
Although our framework shows clear effectiveness in enhancing SPFEM tasks, it still has two main limitations. First, integrating the neighboring AD-THG models with existing SPFEM backbones substantially increases both the parameter size and computational complexity. As shown in Table \ref{Table 7}, when EAT is combined with NED, the number of parameters grows from 84.82M to 288.33M and FLOPs increase from 320.92G to 991.42G. Similarly, DSM exhibits the same trend. Second, the overall generation quality of our framework is ultimately constrained by the inherent limitations of the underlying SPFEM models. Since the integrated AD-THG module primarily serves to refine and temporally align the frames produced by the SPFEM backbone—rather than generating them from scratch—any artifacts, blurriness, or imperfections originating in the backbone cannot be fully eliminated.

To address the limitations discussed above, future work could focus on developing a unified framework that treats expression editing and lip synchronization as distinct yet seamlessly integrated modules. Building on SPFEM, a dedicated lip-synchronization component can be introduced to refine the intermediate representations generated during expression editing. Audio signals are fused with these intermediate features through cross-attention, thereby improving the precision of mouth shape synthesis. These refined features are then injected into multiple layers of the renderer to strengthen their influence on the final output. Unlike the current approach that applies adjacent-frame learning only to AD-THG, this unified framework would enable joint optimization of the expression editor, lip refiner, and renderer under a shared adjacent-frame learning strategy. The integration is expected to yield stronger temporal coherence and higher visual fidelity. Importantly, the modular yet unified design reduces reliance on backbone performance while producing more natural, robust, and realistic results, without incurring significant increases in parameters or computational cost.

\begin{table}[h]
\begin{tabular}{c|c|ccc}
\toprule
Method & Parameters(M) & FLOPs(G) \\
\hline
NED        & 84.82  & 320.92  \\
NED+EAT    & 288.33 & 991.42  \\ 
NED+EAMM   & 197.58 & 597.02  \\
\hline
DSM        & 81.93  & 192.38  \\
DSM+EAT    & 285.44 & 862.88  \\
DSM+EAMM   & 197.69 & 468.48  \\
                 
\bottomrule 
\end{tabular}
\color{black}
\caption{Comparison of model complexity in terms of Parameters and FLOPs for NED and DSM with the integration of EAT or EAMM model.}
\label{Table 7}
\end{table}

\section{Conclusion}
In this work, we propose a framework named THFEM that efficiently integrates AD-THG models and SPFEM models. This framework leverages the robust lip generation capabilities of the AD-THG model, which relies on audio inputs to enhance the naturalness and accuracy of mouth shape during facial expression manipulation. Meanwhile, the AD-THG model is augmented with an adjacent frame learning strategy. Specifically, this strategy first utilizes a single frame generated by the SPFEM model as prior information and feeds the corresponding $n$ frames of audio into the model to predict the $n$ frames image sequences, thereby increasing the authenticity of the generated images. We conduct comprehensive experiments that integrate two advanced AD-THG models with two state-of-the-art SPFEM models. A variety of quantitative and qualitative comparisons, along with user studies, have substantiated the effectiveness and superiority of the proposed framework. 

\begin{acks}
This work was supported in part by the Natural Science Foundation of Guangdong Province under Grant No. 2025A1515010454, and in part by the National Natural Science Foundation of China (NSFC) under Grant No. 62206060.
\end{acks}

\bibliographystyle{ACM-Reference-Format}
\bibliography{ref}

\appendix
\section*{Online Appendix}
\subsection{Additional Analysis of Loss Function Contributions}

To better highlight the contribution of each loss component, we isolate their individual effects via ablations. Since the adversarial loss is indispensable for generative modeling, it is retained in all settings, and the analysis focuses on the remaining three terms. Following the weighting scheme defined in Eq. \ref{eq:loss-function}, we train the five frame neighboring EAT model and apply a controlled variable strategy in which one loss term is removed at a time while the remaining terms are kept fixed. During evaluation, the neighboring EAT model receives a single NED-generated frame together with five consecutive audio frames as input and predicts the subsequent five frames under the intra-identity setting. 

As shown in Table \ref{Table 8}, the model with all losses achieves the best overall performance. Removing the perceptual loss causes the model to fail to converge, highlighting its critical role in preserving global facial structure. Excluding the pixel loss increases FAD from 3.167 to 3.356 and reduces CSIM from 0.874 to 0.856, demonstrating its importance in retaining pixel-level fidelity. Similarly, omitting the sync loss raises LSE-D from 9.415 to 9.698, confirming its necessity for accurate lip-audio synchronization. Overall, these results demonstrate that the joint optimization of perceptual, pixel, and synchronization losses is crucial for ensuring coherent facial structure, fine-grained detail preservation, and natural lip movements.

\begin{table}[H]
\centering
\begin{tabular}{lccc}
\toprule
Setting & FAD↓ & LSE-D↓ & CSIM↑ \\
\midrule
Ours          & 3.167 & 9.415 & 0.874 \\
Ours w/o Perceptual loss & --    & --    & --    \\
Ours w/o pixel loss    & 3.356 & 9.506 & 0.856 \\
Ours w/o Sync loss       & 3.081 & 9.698 & 0.870 \\
\bottomrule
\end{tabular}
\color{black}
\caption{Analysis of the contributions of different loss functions under the intra-identity setting.}
\label{Table 8}
\end{table}

\subsection{Generalization Study on RAVDESS}
To further validate the robustness and generalization of our framework, we perform supplementary experiments on the RAVDESS dataset, a widely adopted benchmark in SPFEM research. We select six speakers (actors 1-6) with 168 videos. Among them, 90\% of the videos are randomly sampled for training, and the remaining 10\% are used for testing.
\newline
\indent
We present the evaluation results of the SPFEM models combined with neighboring AD-THG models on the RAVDESS dataset in Table \ref{Table 9} and Table \ref{Table 10}. In the intra-identity setting, the baseline NED achieves an average FAD of 3.853, CSIM of 0.836, and LSE-D of 8.331. Integrating the neighboring EAT decreases the average FAD and LSE-D to 3.406 and 7.875, respectively, and increases the CSIM to 0.839. When combined with neighboring EAMM, the LSE-D decreases to 8.091 and the CSIM increases to 0.858. Compared to the baseline DSM, which obtains average FAD, CSIM and LSE-D of 0.937, 0.789 and 8.190, integrating EAT decreases FAD and LSE-D to 3.563 and 7.991, and improves the CSIM to 0.806. When combined with EAMM, the LSE-D decreases to 8.189 and the CSIM increases to 0.817. 
In the cross-identity setting, the baseline NED achieves an average FAD, CSIM and LSE-D of 5.402, 0.760 and 8.057. With the integration of EAT, the FAD decreases to 5.126, the LSE-D decreases to 7.874 and the CSIM improves to 0.777. When combined with EAMM, the CSIM increases further to 0.779. Similarly, the baseline DSM obtains average FAD, CSIM and LSE-D of 4.256, 0.756 and 8.165. Incorporating EAT reduces the FAD to 3.879 and the LSE-D to 8.053, while the CSIM improves to 0.775. When combined with EAMM, the CSIM increases slightly to 0.778.
\newline
\indent
In both intra- and cross-identity settings, the evaluation on the RAVDESS dataset shows consistent improvements across the employed evaluation metrics when incorporating the proposed neighboring AD-THG models. These findings confirm the effectiveness of our framework and highlight its strong generalization capability.

\begin{table}[h]
    \scalebox{0.7}{
    \setlength{\tabcolsep}{1pt}
    \begin{tabular}{ 
    c|ccc|ccc|ccc|ccc|ccc|ccc}
    \toprule
    \multirow{2}{*}{Emotions}  &\multicolumn{3}{c|}{DSM}  &\multicolumn{3}{c|}{DSM+EAMM}   &\multicolumn{3}{c|}{DSM+EAT}  &\multicolumn{3}{c|}{NED}  &\multicolumn{3}{c|}{NED+EAMM}  &\multicolumn{3}{c}{NED+EAT}\\
    \cline{2-19}
            & FAD$\downarrow$  & LSE-D$\downarrow$  & CSIM$\uparrow$  & FAD$\downarrow$    & LSE-D$\downarrow$  & CSIM$\uparrow$ & FAD$\downarrow$    & LSE-D$\downarrow$  & CSIM$\uparrow$ & FAD$\downarrow$    & LSE-D$\downarrow$  & CSIM$\uparrow$ & FAD$\downarrow$    & LSE-D$\downarrow$  & CSIM$\uparrow$ & FAD$\downarrow$    & LSE-D$\downarrow$  & CSIM$\uparrow$ \\
    \hline
    Neutral   & 1.632 & 8.083 & 0.880   & 2.790  & 8.249 & 0.879   & 0.833 & 8.134 & 0.903   & 1.619 & 8.148 & 0.901  & 2.727 & 8.281 & 0.892  & 1.383 & 8.104 & 0.892 \\
    Angry     & 4.654 & 8.369 & 0.727   & 4.482  & 8.292 & 0.746   & 4.176 & 8.207 & 0.758   & 4.755 & 8.804 & 0.796  & 4.417 & 8.290 & 0.816  & 6.003 & 7.802 & 0.764 \\
    Disgusted & 6.717 & 8.376 & 0.698   & 4.852  & 8.294 & 0.816   & 6.740 & 7.920 & 0.738   & 6.308 & 8.141 & 0.789  & 5.173 & 8.012 & 0.877  & 3.473 & 7.666 & 0.840 \\
    Fear      & 4.301 & 9.092 & 0.777   & 4.588  & 8.296 & 0.789   & 3.866 & 7.882 & 0.783   & 3.748 & 8.824 & 0.831  & 4.515 & 8.124 & 0.842  & 3.416 & 7.667 & 0.813 \\
    Happy     & 3.574 & 8.201 & 0.835   & 4.217  & 7.989 & 0.844   & 3.280 & 7.914 & 0.836   & 3.570 & 8.814 & 0.875  & 4.498 & 7.846 & 0.876  & 3.410 & 8.053 & 0.887 \\
    Sad       & 2.947 & 8.187 & 0.792   & 3.541  & 8.130 & 0.853   & 3.715 & 7.939 & 0.810   & 3.851 & 8.272 & 0.810  & 3.582 & 7.956 & 0.870  & 3.490 & 7.851 & 0.825 \\
    Surprised & 2.947 & 8.023 & 0.814   & 4.042  & 8.074 & 0.793   & 2.333 & 7.974 & 0.817   & 3.121 & 8.005 & 0.815  & 4.004 & 8.131 & 0.830  & 2.662 & 7.984 & 0.851 \\
    Avg.      & 3.937 & 8.190 & 0.789   & 4.073  & 8.189 & 0.819   & 3.563 & 7.991 & 0.806   & 3.853 & 8.331 & 0.836  & 4.131 & 8.091 & 0.858  & 3.406 & 7.875 & 0.839 \\

    \bottomrule
    \end{tabular}
    \color{black}}
    \caption{Comparison results of  FAD, CSIM, and LES-D of NED, DSM with and without Neighboring AD-THG models in the intra-identity setting on the RAVDESS dataset. }
    \label{Table 9}
\end{table}

\begin{table}[h]
    \scalebox{0.7}{
    \setlength{\tabcolsep}{1pt}
    \begin{tabular}{ 
    c|ccc|ccc|ccc|ccc|ccc|ccc}
    \toprule
    \multirow{2}{*}{Emotions}  &\multicolumn{3}{c|}{DSM}  &\multicolumn{3}{c|}{DSM+EAMM}   &\multicolumn{3}{c|}{DSM+EAT}  &\multicolumn{3}{c|}{NED}  &\multicolumn{3}{c|}{NED+EAMM}  &\multicolumn{3}{c}{NED+EAT}\\
    \cline{2-19}
            & FAD$\downarrow$  & LSE-D$\downarrow$  & CSIM$\uparrow$  & FAD$\downarrow$    & LSE-D$\downarrow$  & CSIM$\uparrow$ & FAD$\downarrow$    & LSE-D$\downarrow$  & CSIM$\uparrow$ & FAD$\downarrow$    & LSE-D$\downarrow$  & CSIM$\uparrow$ & FAD$\downarrow$    & LSE-D$\downarrow$  & CSIM$\uparrow$ & FAD$\downarrow$    & LSE-D$\downarrow$  & CSIM$\uparrow$ \\
    \hline
    Neutral   & 1.885 & 8.108 & 0.848    & 3.353  & 8.374 & 0.845    & 1.885 & 8.240 & 0.876   & 3.559 & 7.906 & 0.820  & 4.663 & 8.225 & 0.803  & 3.073 & 8.050 & 0.838 \\
    Angry     & 4.707 & 8.330 & 0.706    & 4.558  & 8.426 & 0.727    & 4.482 & 8.077 & 0.747   & 5.546 & 8.077 & 0.766  & 5.375 & 8.319 & 0.759  & 5.271 & 8.091 & 0.723 \\
    Disgusted & 6.870 & 8.262 & 0.694    & 4.940  & 8.272 & 0.809    & 6.801 & 7.794 & 0.727   & 7.342 & 8.151 & 0.741  & 6.125 & 8.091 & 0.845  & 6.929 & 7.464 & 0.803 \\
    Fear      & 4.792 & 8.207 & 0.732    & 5.336  & 8.514 & 0.731    & 4.489 & 8.044 & 0.748   & 5.006 & 8.153 & 0.749  & 5.549 & 8.209 & 0.759  & 4.836 & 7.974 & 0.740 \\
    Happy     & 3.818 & 8.259 & 0.801    & 4.694  & 8.404 & 0.805    & 3.503 & 7.977 & 0.800   & 5.644 & 8.087 & 0.804  & 6.415 & 7.836 & 0.820  & 5.787 & 7.932 & 0.833 \\
    Sad       & 4.426 & 7.980 & 0.745    & 4.546  & 8.276 & 0.787    & 4.395 & 8.060 & 0.760   & 5.583 & 8.034 & 0.726  & 5.417 & 8.106 & 0.769  & 5.370 & 7.767 & 0.765 \\
    Surprised & 3.292 & 8.008 & 0.763    & 4.940  & 8.336 & 0.741    & 2.727 & 8.177 & 0.770   & 5.135 & 7.993 & 0.716  & 6.187 & 8.348 & 0.697  & 4.614 & 7.830 & 0.736 \\
    Avg.      & 4.256 & 8.165 & 0.756    & 4.624  & 8.372 & 0.778    & 3.879 & 8.053 & 0.775   & 5.402 & 8.057 & 0.760  & 5.675 & 8.162 & 0.779  & 5.126 & 7.874 & 0.777 \\

    \bottomrule
    \end{tabular}
    \color{black}}
    \caption{Comparison results of  FAD, CSIM, and LES-D of NED, DSM with and without Neighboring AD-THG models in the cross-identity setting on the RAVDESS dataset. }
    \label{Table 10}
\end{table}

\end{document}